\documentclass[twocolumn,10pt]{IEEEtran}
\renewcommand{\baselinestretch}{0.969}
\usepackage{amsmath, graphics,amssymb,epsfig}
\usepackage{cite}
\usepackage{url}
\usepackage{color,soul,xcolor}
\usepackage{hhline}
\usepackage{bbold}
\usepackage{makecell}
\usepackage{amsthm}
\usepackage{kotex}
\usepackage{caption}
\usepackage{subcaption}
\usepackage[font=footnotesize,labelsep=period]{caption}
\def\BibTeX{{\rm B\kern-.05em{\sc i\kern-.025em b}\kern-.08em
    T\kern-.1667em\lower.7ex\hbox{E}\kern-.125emX}}

\newtheorem{prop}{Proposition}

\newfont{\bb}{msbm10 scaled 1000}

\newcommand{\be}{\begin{equation}}
\newcommand{\ee}{\end{equation}}
\newcommand{\bea}{\begin{eqnarray}}
\newcommand{\eea}{\end{eqnarray}}
\newcommand{\bitem}{\begin{itemize}}
\newcommand{\eitem}{\end{itemize}}

\makeatletter
\def\hlinewd#1{%
\noalign{\ifnum0=`}\fi\hrule \@height #1 \futurelet \reserved@a\@xhline}

\makeatother

\makeatletter

\newcommand*{\rom}[1]{\expandafter\@slowromancap\romannumeral #1@}
\makeatother

\begin{document}

\setlength{\abovedisplayskip}{5pt}
\setlength{\belowdisplayskip}{5pt}

\title{{\LARGE Deep Learning Based Resource Assignment for Wireless Networks}}
\vspace{-1mm}

\author{\IEEEauthorblockN{Minseok Kim, Hoon Lee*, \textit{Member}, \textit{IEEE}, Hongju Lee, and Inkyu Lee, \textit{Fellow}, \textit{IEEE}  \vspace*{-9mm}}
\thanks{This work was supported in part by the National Research Foundation of Korea (NRF) grant funded by the Korea Government (MSIT) under Grant 2017R1A2B3012316
\textit{(Corresponding authors : Inkyu Lee and Hoon Lee)}}
\thanks{M. Kim, H. Lee, and I. Lee are with the School of Electrical Engineering, Korea University, Seoul, Korea (e-mail: $\{$msk1005, honglee2335, inkyu$\}$@korea.ac.kr).}
\thanks{*H. Lee is with the Department of Smart Robot Convergence and Application Engineering and the Department of Information and Communications Engineering, Pukyong National University, Busan 48513, Korea (e-mail: hlee@pknu.ac.kr).}
}\maketitle

\begin{abstract}
This paper studies a deep learning approach for binary assignment problems in wireless networks, which identifies binary variables for permutation matrices. This poses challenges in designing a structure of a neural network and its training strategies for generating feasible assignment solutions. To this end, this paper develop a new Sinkhorn neural network which learns a non-convex projection task onto a set of permutation matrices. An unsupervised training algorithm is proposed where the Sinkhorn neural network can be applied to network assignment problems. Numerical results demonstrate the effectiveness of the proposed method in various network scenarios.
\end{abstract}
\vspace*{-1.5mm}
\vspace{-6mm}
\section{Introduction} \label{sec:introduction} 
\vspace*{-1mm}
Assignment problems which determine matching between two different quantities have been prevailed in various networking scenarios.
Popular examples are subcarrier allocation \cite{HJKIM:16}, user-cell association \cite{cell:20}, and task offloading \cite{JCN:20}. Several algorithms have been proposed for handling such assignment problems. In particular, the Hungarian algorithm \cite{Hungarian:1955} was introduced as a globally optimal solver for linear sum assignment problems (LSAPs). Mixed integer programs (MIPs) solvers, e.g., Mosek and Gurobi \cite{Gurobi}, can address convex assignment problems. However, the technique typically invokes special properties for objective functions such as linearity and convexity, and involves high computational complexity for executing iterative calculations.

Recently, deep learning (DL) based optimization methods have been adopted as promising tools in wireless networks for reducing the computational burden of traditional iterative algorithms \cite{LtoO:18}. A supervised learning approach which trains deep neural network (DNNs) to memorize solutions of existing optimization techniques has been investigated for power control problems \cite{LtoO:18}. The DNNs achieve near-optimal performance with reduced execution time. However, the supervised learning invokes high complexity for a data collection step for securing numerous known optimal solutions for training. Therefore, it can be applicable to simple networking scenarios where efficient optimization algorithms are available. To resolve this issue, an unsupervised learning concept \cite{DPC:18} was proposed where DNNs are designed to model optimal solution computation rules. 
The effectiveness of the unsupervised DL strategy has been verified for optimizing non-convex problems in various wireless systems \cite{HLee:19b}. Although the unsupervised DL does not require the information of the optimal computation strategy.

There have been lots of efforts on the development of low-complexity DL methods for assignment problems. In particular, the LSAP which formulates a minimization task of the linear network cost function has been recently solved via the supervised DL approach \cite{LSAP:18}. An assignment problem was decomposed into several sub-assignment problems, which are regarded as classification tasks that find one-way matching of a certain item. Individual DNNs are dedicated to solving each subproblem. The optimum solution generated by the Hungarian algorithm facilitates supervised learning of multiple DNNs simultaneously.
The combinatorial nature of the assignment problems calls for a feasible solution to be structured in a permutation matrix format. However, such a strict feasibility condition would not be guaranteed by the conventional supervised DL technique \cite{LSAP:18}. To handle this issue, a post-processing method can be adopted to recover permutation matrices, but it results in performance loss of the trained DNN. This challenge poses in existing unsupervised DL-based optimization approaches since they are confined to the optimization of continuous-valued variables, but not permutation matrices. Therefore, it is necessary to develop a new learning structure that directly identifies feasible assignment solution as an output of a DNN.

This paper proposes an unsupervised DL framework for generic non-convex assignment problems. Existing convex MIP solvers can only address a certain instance of assignment tasks. On the contrary, this paper aim at identifying an efficient mapping, i.e., a DNN, that generates assignment solutions for arbitrary problem instances. As a result, it can be applied to general non-convex assignment problems. The major challenge is to construct a DNN which always generates feasible solutions for arbitrary network assignment tasks. To overcome this difficulty, this paper introduce a novel Sinkhorn neural network (SNN). The output layer of the SNN is designed to solve non-convex projection problems onto the set of permutation matrices, thereby ensuring the feasibility as assignment problem solvers. The SNN is trained in an unsupervised manner without knowing the optimum solutions. Numerical results validate the efficacy of the proposed DL methods in practical network assignment scenarios. It is verified that the proposed SNN approach achieves almost identical performance to existing algorithms with reduced complexity.
\vspace{-2mm}



\vspace*{-3mm}
\section{Problem Description} \label{sec:Problem Statement}
\vspace*{-1.5mm}
Consider network assignment problems which determine optimal matching policies between two distinct wireless nodes, in particular, bipartite matching from $M$ nodes in $\mathcal{M}=\{1,\cdots,M\}$ to $N$ nodes $\mathcal{N}=\{1,\cdots,N\}$. This can be interpreted as associations among base stations (BSs), user equipments (UEs), and resource blocks. The balanced case $M=N$ is discussed first and it will be extended to a general unbalanced case of $N>M$ later. The target of the balanced assignment problem with $N=M$ is to identify an one-to-one allocation strategy among nodes. 

Let $x_{ij}\in\{0,1\}$ as a binary variable indicating the assignment state of nodes $i\in\mathcal{M}$ and $j\in\mathcal{N}$, i.e., $x_{ij}=1$ if entity $i$ is assigned to entity $j$ and $x_{ij}=0$ otherwise. Each node can connect to only one node, which imposes the constraints as
\begin{align}
\sum\nolimits_{i=1}^{N}x_{ij} = 1,\forall j,\ \text{and } \sum\nolimits_{j=1}^{N}x_{ij}  = 1, \forall i, \label{eq:unitsum}
\end{align}
where $x_{ij}$ collectively form a binary assignment matrix $\mathbf{X} \triangleq \{x_{ij},\forall i,j\} \in \{0,1\}^{N\times N}$. It is inferred from \eqref{eq:unitsum} that a feasible assignment solution $\mathbf{X}$ should be a permutation matrix. 

Let $\mathbf{H} \triangleq \{h_{ij},\forall i,j\} \in \mathbb{R}^{N\times N}$ be the input of the assignment problem describing wireless propagation environments. In particular, $h_{ij}$ represents the connection status between nodes $i$ and $j$, e.g. channel coefficients, which determine the cost of the assignment $x_{ij}$. A generic network assignment problem can be described as a minimization of the network cost function $f(\mathbf{X},\mathbf{H})$, possibly non-convex, subject to the constraint on $\mathbf{X}$ being a permutation matrix. It is written by
\begin{align}
&\min_{\mathbf{X}\in\mathcal{P}_{N}} \ f(\mathbf{X},\mathbf{H}), 
\label{eq:assignP}
\end{align}
where $\mathcal{P}_{N}$ is the set of all $N$-by-$N$ permutation matrices.

The assignment problem (\ref{eq:assignP}) prevails in a design of wireless networks. Existing convex MIP solvers such as the branch-and-bound algorithm \cite{B&B:1977} cannot handle the non-convex cost $f(\mathbf{X},\mathbf{H})$. The Hungarian algorithm has been known as the optimum algorithm for the LSAP, which is a special case of (\ref{eq:assignP}) with the affine cost function $f(\mathbf{X},\mathbf{H})=\text{tr}(\mathbf{H}^{T}\mathbf{X})$, but it cannot address generic non-convex cost functions. The dual function of (\ref{eq:assignP}) would not be available due to the non-convex cost function, thereby making the Lagrange duality method intractable.

In this paper, we propose a DL approach to solve the generic non-convex assignment problem \eqref{eq:assignP} that can be applicable to various networking setups. A key idea is to exploit the ``learning to optimize'' concept \cite{LtoO:18} which replaces unknown optimization processes with properly trained DNNs. The optimization procedure of (\ref{eq:assignP}) can be viewed as an identification task of a mapping from the network states $\mathbf{H}$ to the permutation matrix $\mathbf{X}$. Such a mapping is implemented by a $R$-layer fully-connected DNN $\mathcal{F}(\cdot;\theta)$ with a parameter set $\theta$. 

The input and the output of the DNN are modeled $\mathbf{h}\triangleq\text{vec}(\mathbf{H})$ and $\mathbf{x}\triangleq\text{vec}(\mathbf{X})$, respectively, where $\text{vec}(\cdot)$ represents vectorization of a matrix. Denoting $K_r$ as the dimension of the $r$-th layer, the computation of the DNN is expressed as
\vspace*{-1mm}
\bea
\mathcal{F}(\mathbf{h};\theta) = \sigma_R(\mathbf{W}_R \times \cdots \times \sigma_1(\mathbf{W}_1\mathbf{h} + b_1) + \cdots + \mathbf{b}_{R})\label{eq:forward}
\eea
where an element-wise function $\sigma_r(\cdot)$ is an activation function at layer $r$ $(r = 1,\cdots,R)$ and $\mathbf{W}_r \in \mathbb{R}^{K_{r}\times K_{r-1}}$ and $\mathbf{b}_r \in \mathbb{R}^{K_{r}}$ indicate the weight matrices and the bias vectors, respectively. Here $\mathbf{W}_r \text{ and } \mathbf{b}_r$ collectively form the trainable parameter set $\theta\triangleq\{\mathbf{W}_{r},\mathbf{b}_{r},\forall r\}$.
In the training step, the parameter $\theta$ is identified such that the DNN output $\mathbf{x}=\mathcal{F}(\mathbf{h};\theta)$ minimizes the cost function $f(\mathbf{X},\mathbf{H})$ in (\ref{eq:assignP}) for a given network state $\mathbf{H}$ while satisfying the permutation constraint $\mathbf{X}\in\mathcal{P}_{N}$. However, since conventional training algorithms were developed for unconstrained formulations, the feasibility of the DNN output cannot be guaranteed. 

The recent work \cite{LSAP:18} presented a DL method for tackling the LSAPs. To ensure the feasibility, a supervised learning strategy was adopted which forces the DNN to yield the optimal permutation matrix obtained by the Hungarian algorithm. Nevertheless, the DNN would fail to generate feasible assignment matrices for unseen test samples. This leads to unintended collisions in matching between jobs and workers. Thus, a post-processing was added in the test step which reassigns conflicting jobs to a worker with the lowest cost value in a heuristic way. Although the feasibility may be secured, it might incur a loss of the optimality since the post-processing was not involved in the training step. Furthermore, due to the supervised learning concept, the method in \cite{LSAP:18} can only be applicable to simple assignment problems having efficient solvers. Therefore, it is necessary to develop a new DNN structure which is able to capture the non-convex constraint of generic assignment tasks with arbitrary network cost functions.
\vspace{-7mm}
\section{Proposed Deep Learning Approach}\label{sec:sec3}
\vspace{-1mm}
This section proposes a SNN for solving the assignment problem (\ref{eq:assignP}). It is desired to determine the output activation $\sigma_{R}(\cdot)$ of the DNN $\mathcal{F}(\cdot;\theta)$ which always generates proper permutation matrices for any given inputs $\mathbf{H}$. To this end, a novel activation function is developed to carry out non-convex projections onto the set of permutation matrices $\mathcal{P}_{N}$. 
Let $\mathbf{a}\in\mathbb{R}^{N^{2}}$ be the output vector of the DNN with given input $\mathbf{H}$. Then, $\mathbf{a}$ is the input vector to the output activation $\sigma_{R}(\cdot)$ at the end of the DNN.

Firstly, $\mathbf{a}$ is reshaped into an $N$-by-$N$ matrix $\mathbf{A}$, and then pass to the output activation $\mathbf{X}=\sigma_{R}(\mathbf{A})$ which solves the non-convex projection problem as 
\vspace*{-1mm}
\begin{align}
    \sigma_{R}(\mathbf{A})\triangleq\arg\max_{\mathbf{X}\in\mathcal{P}_{N}}\text{tr}(\mathbf{A}^{T}\mathbf{X}).\label{eq:proj}
    \vspace*{-1mm}
\end{align}
Problem \eqref{eq:proj} determines a permutation matrix that maximizes the affinity between the output feature $\mathbf{A}$. The output activation \eqref{eq:proj} can always satisfy the conditions $\eqref{eq:unitsum}$ for an arbitrary input $\mathbf{H}$. However, the combinatorial nature of \eqref{eq:proj} invokes a selection process which nullifies the gradient with respect to $\mathbf{A}$, posing challenges for gradient-based training algorithms. 

To address this issue, a soft approximation of the non-convex projection \eqref{eq:proj} is introduced. The key idea is to exploit the concept of the Sinkhorn operation \cite{Mena:18} which has been originally designed for obtaining a doubly stochastic matrix (DSM). The DSM $\mathbf{D}\in[0,1]^{N\times N}$ is defined as a square matrix whose $(i,j)$-th elements $d_{ij}$ satisfy the constraint in \eqref{eq:unitsum} as $\sum_{i=1}^{N}d_{ij}=1$ and $\sum_{j=1}^{N}d_{ij}=1$ with $0 \leq d_{ij} \leq 1$.

Thus, the DSM can be regarded as a continuous relaxation of a permutation matrix. The Sinkhorn operator denoted by $S(\mathbf{A})$ calculates a projection of $\mathbf{A}$ into a convex set containing DSMs. The corresponding problem can be written as
\vspace*{-1mm}
\begin{align}
    S(\mathbf{A})=&\arg\max_{\mathbf{D}\in\mathcal{D}_{N}}\text{tr}(\mathbf{A}^{T}\mathbf{D})\label{eq:SA},
    \vspace*{-1mm}
\end{align}
where $\mathcal{D}_{N}$ stands for the set of $N$-by-$N$ DSMs.

A solution of \eqref{eq:SA} can be found by iteratively normalizing rows and columns of $\mathbf{A}$. The row-wise normalization $R(\mathbf{A})\triangleq\{\widetilde{a}_{ij},\forall i,j\}$ and the column-wise normalization $C(\mathbf{A})\triangleq\{\widehat{a}_{ij},\forall i,j\}$ are respectively given with
\vspace*{-1mm}
\bea
    \widetilde{a}_{ij}=\frac{a_{ij}}{\sum_{k=1}^{N}a_{ik}}\ \text{and}\ \widehat{a}_{ij}=\frac{a_{ij}}{\sum_{k=1}^{N}a_{kj}},  \label{eq:normalization}
\eea
where $a_{ij}$ is the $(i,j)$-th element of $\mathbf{A}$.

Then, the computation of the Sinkhorn operator at the $m$-th iteration $S^{m}(\mathbf{A})$ is written by
\vspace*{-1mm}
\begin{align}
    S^{m}(\mathbf{A})=C(R(S^{m-1}(\mathbf{A}))),\label{eq:cSinkhorn}
\end{align}
where $S^{0}(\mathbf{A})\triangleq\exp(\mathbf{A})$ denotes an initial Sinkhorn operator with $\exp(\cdot)$ being an element-wise exponential function. 
It has been revealed in \cite{Mena:18} that iterating \eqref{eq:cSinkhorn} converges to the optimal point of \eqref{eq:SA}, i.e., $S(\mathbf{A})\triangleq\lim_{m\rightarrow\infty}S^{m}(\mathbf{A})$. 
In the following proposition, the relationship between the Sinkhorn operator $S(\mathbf{A})$ and permutation matrices is provided.
\vspace{-2mm}
\begin{prop}\label{prop:prop1}
Suppose that elements of $\mathbf{A}\in\mathbb{R}^{N\times N}$ are independent and their distributions are absolutely continuous in the Lebesgue measure. Then, $\lim_{\tau\to\infty}S(\tau\mathbf{A})$ almost surely provides a permutation matrix.
\end{prop}
\vspace{-5mm}
\begin{proof}
The Birkhoff theorem \cite{Birkhoff:1946} states that any DSM is given as a convex combination of permutation matrices, i.e., the convex hull of $\mathcal{P}_{N}$ becomes a polytope $\mathcal{D}_{N}$ generated by DSMs. Therefore, vertices of $\mathcal{D}_{N}$ form permutation matrices. It has been verified from the Lagrange duality method that in the extreme case $\tau\rightarrow \infty$, $S(\tau\mathbf{A})$ converges to a vertex of a feasible space $\mathcal{D}_{N}$, i.e., a permutation matrix. Please refer to \cite[Theorem 1]{Mena:18} for the detailed proof.
\end{proof}
\vspace{-2mm}
Proposition \ref{prop:prop1} implies that with a sufficiently large $\tau$, the Sinkhorn operator $S(\tau\mathbf{A})$ can identify a permutation matrix nearest to an arbitrary square matrix $\mathbf{A}$. With the initialization $S^{0}(\tau\mathbf{A})=\exp(\tau\mathbf{A})$, the normalization in \eqref{eq:normalization} is interpreted as a scaled softmax function defined as
\vspace*{-1mm}
\begin{align}
    \text{softmax}(i,z_{j},\tau) = \frac{\text{exp}(\tau z_i)}{\sum_{j\neq i}\text{exp}(\tau z_j)}. \label{eq:softmax}
\end{align}
As $\tau$ gets larger, \eqref{eq:softmax} approaches the argmax function producing an one-hot vector, which is an all-zero vector except for the maximum index being replaced by one. The Sinkhorn operator repeatedly applies the scaled softmax \eqref{eq:softmax} to each row and column of $\mathbf{A}$ so that the output becomes a permutation matrix whose rows and columns have a single one with zeros elsewhere. Consequently, the Sinkhorn operator $S(\tau\mathbf{A})$ with a large $\tau$ solves the non-convex problem \eqref{eq:proj} efficiently. 

Based on these observations, the output activation function $\mathbf{\sigma}_{R}(\cdot)$ of the DNN $\mathcal{F}(\cdot;\theta)$ is designed as the Sinkhorn operator $\mathbf{\sigma}_{R}(\mathbf{A})=S(\tau{\mathbf{A}})$. Thanks to the continuous-valued computations \eqref{eq:normalization}, the Sinkhorn operator $S(\tau\mathbf{A})$ has valid gradients, meaning that existing gradient-based training algorithms can be applied to optimize the SNN parameter $\theta$. 
The parameter $\tau$ controls the quality of the approximation $S(\tau \mathbf{A})\simeq\mathbf{X}\in\mathcal{P}_{N}$. A large $\tau$ leads to a high approximation accuracy. However, the gradient of $S(\tau \mathbf{A})$ may explode as $\tau$ gets larger, thereby prohibiting an efficient training of the DNN via gradient-based algorithms. We thus need to choose $\tau$ carefully through a validation process to achieve a good tradeoff between an approximation accuracy and training performance.
\vspace{-4mm}
\subsection{Training and Implementation}\label{sec:sec3a}
\vspace{-1mm}
A training strategy is presented for the proposed SNN. By substituting $\mathbf{X}=\mathcal{F}(\mathbf{h};\theta)$ into (\ref{eq:assignP}), it follows
\vspace*{-1mm}
\begin{align}
\min_{\theta} \ \ f(\mathcal{F}(\mathbf{h};\theta),\mathbf{H}) \label{eq:train}
\end{align}
where the constraint $\mathcal{F}(\mathbf{h};\theta)\in\mathcal{P}_{N}$ can be ignored since the SNN always generates permutation matrices. Compared to the original formulation (\ref{eq:assignP}), the optimization variable $\mathbf{X}$ now turn out to be the SNN parameter $\theta$. Hence, \eqref{eq:train} becomes a training task which determines an efficient SNN for handling assignment problems with an arbitrary network observation $\mathbf{H}$.

To solve \eqref{eq:train}, a training dataset $\mathcal{T}\triangleq\{\mathbf{H}\}$ containing numerous realizations of $\mathbf{H}$ is first prepared. Then, the SNN is trained to minimize the cost function averaged over the training dataset. This can be solved by gradient-based learning algorithms, e.g., the mini-batch stochastic gradient descent (SGD) \cite{SGD:15}, which iteratively updates $\theta$ for minimizing the cost function evaluated over a sample dataset called the mini-batch set. The SNN parameter $\theta^{[t]}$ at the $t$-th training epoch of the SGD algorithm is calculated~as
\bea
    \theta^{[t]} = \theta^{[t-1]} - \frac{\eta}{|\mathcal{B}|}\sum\nolimits_{\mathbf{H}\in\mathcal{B}}\nabla f(\mathcal{F}(\mathbf{h};\theta),\mathbf{H})\label{eq:sgd}
\eea
where $\eta$ stands for the learning rate, $\mathcal{B}\subset\mathcal{T}$ is the mini-batch set, and $\nabla$ indicates the gradient operator.

The training algorithm in \eqref{eq:sgd} can be implemented in an unsupervised manner without the knowledge of the optimal solution of the original assignment problem (\ref{eq:assignP}). Notice that the conventional supervised DL approach \cite{LSAP:18} needs to collect the optimal assignment matrices, and thus it can only be applied to the LSAP where efficient algorithms for (\ref{eq:assignP}) are available. In contrast, the proposed unsupervised DL framework \eqref{eq:sgd} enables the SNN to learn arbitrary assignment tasks. The trained SNN parameter $\theta$ is realized in any computational unit, e.g., BSs. Then, a solution to an unseen input $\mathbf{H}$ can be obtained by linear calculations~\eqref{eq:forward}. The complexity is given by $\mathcal{O}(\sum_{r=1}^{R}\!K_{r}K_{r\!-\!1})$ where $K_{0}\!\!=\!\!K_{R}\!\!=\!\!N^{2}$ is the length of the input $\mathbf{h}$ and the output $\mathbf{x}$. When the hidden dimensions 
are independent of $N$, the corresponding complexity becomes $\mathcal{O}(N^{2})$. Assuming $L$ iterations of the Sinkhorn operations \eqref{eq:cSinkhorn}, the overall complexity equals $\mathcal{O}(LN^{2})$, which is lower than that of the Hungarian algorithm given by $\mathcal{O}(N^{3})$.



\begin{figure}
     \centering
     \begin{subfigure}{0.2409\textwidth}
         \centering
         \includegraphics[width=\textwidth]{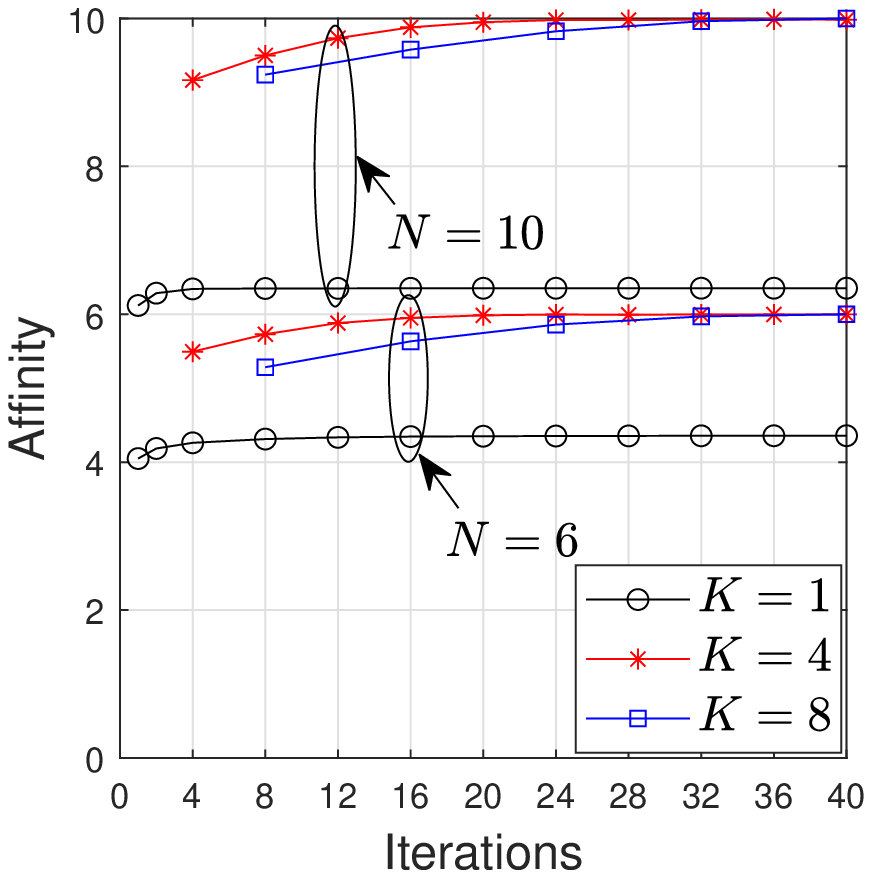}
         \vspace{-6mm}
         \caption{$\tau=20$}
         \label{figure:convergence_a}
     \end{subfigure}
     \hfill
     \begin{subfigure}{0.2409\textwidth}
         \centering
         \includegraphics[width=\textwidth]{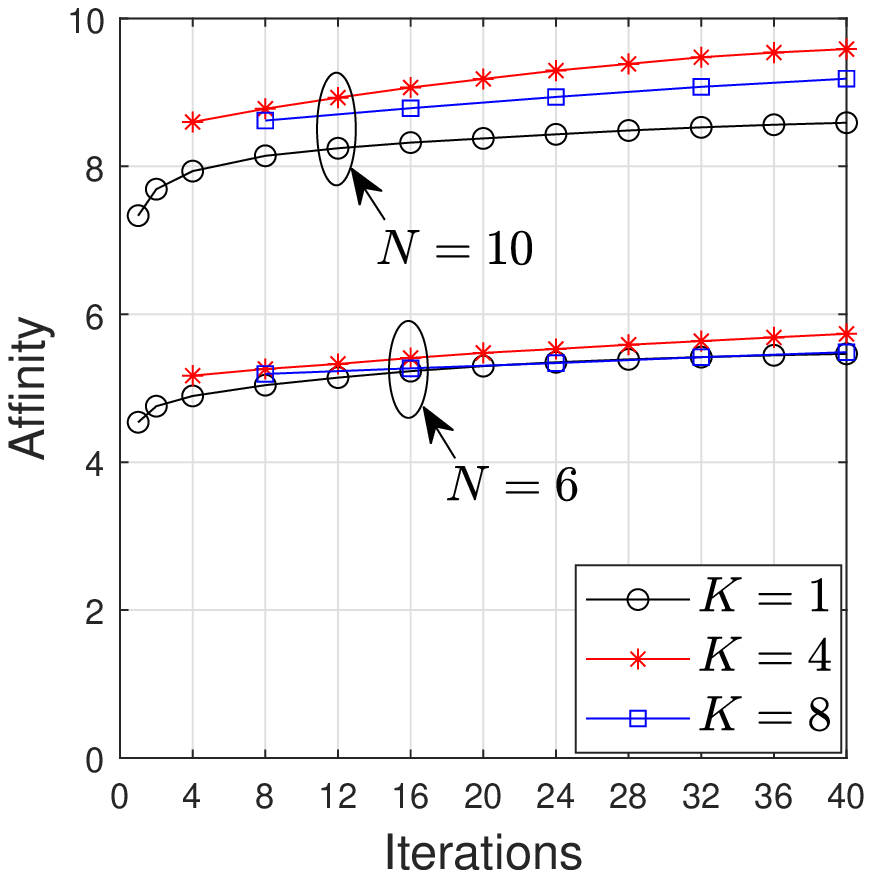}
         \vspace{-6mm}
         \caption{$\tau=100$}
         \label{figure:convergence_b}
     \end{subfigure}
    \vspace{-2mm}
    \caption{Convergence behavior of the output activation for various $K$ and $\tau$.}
    \vspace{-7mm}
    \label{figure:convergence}
\end{figure}

In practice, the Sinkhorn operator invokes a large number of iterations to find an exact permutation matrix. To improve the convergence speed, $K$ consecutive Sinkhorn operators are applied for constructing the output activation $\sigma_{R}(\cdot)$, where each Sinkhorn operator lasts $\frac{L}{K}$ iterations, resulting in total $L$ iterations. Fig. \ref{figure:convergence} exhibits the convergence trends of the output activation for various $K$ and $\tau$ with $N\!=\!6$ and $10$. The evaluation metric is defined as the affinity to the nearest permutation matrix $\max_{\mathbf{X}\in\mathcal{P}_{N}}\text{tr}(S(\tau \mathbf{A})^{T}\mathbf{X})$. Achieving the upperbound value $N$ indicates that an exact permutation matrix is found. A single Sinkhorn operator cannot provide an exact permutation matrix. On the contrary, a cascaded integration of $K\!=\!4$ Sinkhorn operators solves the non-convex projection \eqref{eq:proj} within 20 iterations. This implies that the output activation with multiple Sinkhorn operators significantly enhance the approximation accuracy of the non-convex projection problem \eqref{eq:proj}. Increasing $K$ first helps the output activation converge quickly, but adopting too many Sinkhorn operators may degrade the convergence speed. From the figure, it is concluded that $K\!=\!4$ is an efficient choice for all simulated $N$. A large $\tau$ can improve the feasibility for the constraint $\mathbf{X}\!\in\!\mathcal{P}_{N}$. However it incurs the exploding gradient issue, thereby leading to slow convergence. It has been found that $\tau\!=\!20$ achieves a good tradeoff between the feasibility and the convergence speed.

\vspace{-4mm}
\subsection{Extension to the Unbalanced Case}
\vspace*{-1mm}
Now, the proposed SNN approach is extended for the unbalanced assignment problems with $N>M$. Job $j$ $(j=1,\cdots,M)$ should be allocated to one of $N$ workers, and worker $i$ $(i=1,\cdots,N)$ can handle at most one job. Since $N>M$, some workers may not have a job. 
The assignment matrix $\mathbf{X}\triangleq\{x_{ij},\forall i,j\}\in\{0,1\}^{N\times M}$ and the cost matrix $\mathbf{H}\triangleq\{h_{ij},\forall i,j\}\in\mathbb{R}^{N\times M}$ now become non-square matrices. 
In this configuration, the constraints in \eqref{eq:unitsum} are refined as
\begin{align}
\sum\nolimits_{i=1}^{N}x_{ij}  = 1, \forall j,\ \text{and}\ \sum\nolimits_{j=1}^{M}x_{ij}  \leq 1, \forall i.\label{eq:m-unit-sum}
\end{align}

The proposed SNN method can tackle the unbalanced assignment constraints \eqref{eq:m-unit-sum} with simple modifications. The SNN takes a non-square cost matrix as an input feature and produces a square output matrix denoted by $\tilde{\mathbf{X}}\in\{0,1\}^{N\times N}$. Since each column of the permutation matrix $\tilde{\mathbf{X}}$ have a single one, removing arbitrary $N-M$ columns of $\tilde{\mathbf{X}}$ makes the resultant non-square matrix of size $N$-by-$M$ feasible for \eqref{eq:m-unit-sum}. For simplicity, the last $N-M$ columns of $\tilde{\mathbf{X}}$ are discarded, and the corresponding matrix acts as the non-square assignment solution $\mathbf{X}$. During the training, the cost function $f(\mathbf{H},\mathbf{X})$ is evaluated with the modified non-square assignment matrix $\mathbf{X}$. The associated backpropagation procedure becomes inactive for the discarded variables. As a result, the proposed SNN method can be readily applied to solve the unbalanced cases.

\vspace{-3mm}
\section{Numerical Results} \label{sec:Numerical Results and conclusion}
\vspace*{-1mm}
This section examines the effectiveness of the proposed DL method in various networking scenarios.
The rectified linear unit (ReLU) activation is employed for all hidden layers. The hyperparameters are given as $\tau = 20$, $K = 4$, and $L=20$. The learning rate and the mini-batch size are fixed as $\eta = 10^{-3}$ and $|\mathcal{B}| = 2000$, respectively. The number of training iterations is set to $10^6$, resulting in total $2 \times 10^9$ training samples. The SNN parameter with the minimum cost over $10^4$ validation samples is chosen as the best model. Finally, the performance of the trained SNN is examined over $10^4$ test samples. 

\vspace{-4mm}
\subsection{Linear Sum Assignment Problem}
\vspace*{-1mm}
The LSAP is considered first which can be optimally solved via the Hungarian algorithm \cite{Hungarian:1955}. In this example, we examine the performance the proposed SNN with the optimal solution and the conventional DL method \cite{LSAP:18} in the balanced and unbalanced cases. The SNN is constructed with 3 hidden layers each of which has the dimension 288, 144 and 80, respectively. Elements of the cost matrix $\mathbf{H}$ are uniformly distributed within $[1,100]$. 

\begin{table}[b]
\vspace{-2mm}
\centering
\caption{Average degradation of cost function.}
\vspace{-2mm}
\begin{tabular}{|c||c|c|c|c|}
\hline
$(M,N)$  & \makecell{(4,4)} & \makecell{(8,8)} & \makecell{(2,4)} & \makecell{(4,8)}   \\ \cline{1-5}
\hhline{|=||=|=|=|=|}
Proposed & 0.27\% & 0.85\%   & 0.17\%  & 0.62\%   \\ \cline{1-5}
\hline
Supervised\cite{LSAP:18} &  1.71\% & 22.65\%  & 1.78\%  & 23.23\% \\ \cline{1-5}
\hline
\end{tabular}
\vspace{-5mm}
\end{table}

Table I presents the average degradation of the cost value compared to the optimal Hungarian algorithm in the balanced and unbalanced cases. The system with matching tasks from $M$ nodes to $N$ nodes is denoted by $(M,N)$. For comparison, the performance of the supervised DL approach \cite{LSAP:18} is also evaluated.
A performance loss of the proposed SNN is less then $1\ \%$ for all configurations, whereas the supervised DL exhibits a high loss especially for a large $N$. This stems from the heuristic post-processing steps in \cite{LSAP:18} that recover the structure of permutation matrices. The unbalanced case shows similar performance to that of the balanced one with the same $N$. 
This is because the unbalanced cases are tackled by producing a square output matrix of size $N$-by-$N$. The number of floating point operations of the proposed SNN is obtained as $816N^2 - 20N + 105984$, which is typically lower than that of the learning method in \cite{LSAP:18} which is given by $64N^3 + 512N^2 + 18432N$. In addition, at $N=8$, the CPU execution time of the proposed scheme is 65 times faster than that of the Hungarian algorithm. Thus, it is concluded that the proposed SNN-based unsupervised DL framework is more suitable for handling the LSAP both in terms of the performance and the computational complexity.

\vspace{-4mm}
\subsection{Cell association problem}
\vspace*{-1mm}
Consider a more practical scenario where $N$ single antenna BSs communicate with $N$ single antenna UEs, and the BSs are assumed to support only one UE. A joint optimization of the transmit power at BSs and cell association among BSs and UEs is addressed. The association state between BS $i$ and UE $j$ is denoted by a binary variable $x_{ij}\in\{0,1\}$, whereas the transmit power of BS $i$ is expressed as a continuous variable $p_{i}\in[0,P_{i}]$ with $P_{i}$ being the maximum power budget. Let $h_{ij}$ be the channel gain from BS $i$ to UE $j$. When UE $j$ is supported by BS $i$, the achievable rate is written by
\vspace*{-1mm}
\begin{figure}[t]
\begin{center}
\includegraphics[width=2.8in]{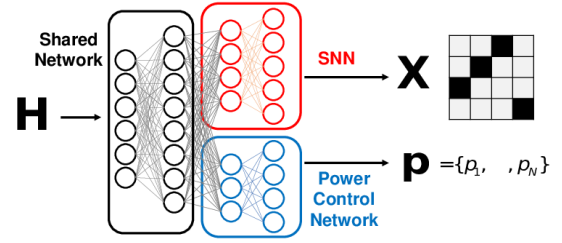}
\end{center}
\vspace{-3mm}
\caption{DNN structure for cell association problem.}
\label{figure:MTL}\vspace*{-5mm}
\end{figure}
\begin{align}
    r_{ij}(\mathbf{p}) =  \log\left(1+\frac{p_{i}h_{ij}}{\sigma^2+\sum_{k\neq i}p_{k}h_{kj}}\right), 
\end{align}
where $\mathbf{p}\triangleq\{p_{j},\forall j\}$ and $\sigma^2$ accounts for the noise power. The sum rate maximization problem is then formulated as
\vspace*{-1mm}
\begin{align}
\max_{\mathbf{X}\in\mathcal{P}_{N},\mathbf{p}} \sum\nolimits_{j=1}^{N}\sum\nolimits_{i=1}^{N}x_{ij}r_{ij}(\mathbf{p}), 
\ s.t.\ p_{i}\in[0,P_{i}], \forall i. \label{eq:power}  
\end{align}
The non-convex MIP in \eqref{eq:power} jointly optimizes continuous-valued power variables $p_{i}$ and binary assignment variables $x_{ij}$.


As illustrated in Fig. \ref{figure:MTL}, a DNN for tackling \eqref{eq:power} is constructed with three different neural networks. The channel matrix $\mathbf{H}\triangleq\{h_{ij},\forall i,j\}$ is first processed by a shared network, and its output is fed to a SNN and another DNN for power control. The SNN identifies a cell association matrix $\mathbf{X}$, while the power control network calculates the power allocation solution $\mathbf{p}$. The sigmoid activation is adopted at the output layer for satisfying the power budget constraint. 

For evaluation purposes, a two-tier heterogeneous network is considered where BS $1$ is regarded as a macro-cell BS located at the center cell of radius $1$ km. The remaining $N\!-\!1$ BSs act as small-cell BSs that encircle the macro-BS with radius $500$ m. UEs are randomly deployed with the minimum distance $10$ m from BSs. We adopt the path-loss model $120.9\!+\!37.6\log_{10}\!d$ dB with $d$ being the distance from a BS and an UE. The Rayleigh fading is considered for the small-scale channel gains. The standard deviation of the log-normal shadowing is fixed as $8$ dB, and the noise power is set to $\sigma^{2}\!=\!-\!114$ dBm. The shared network consists of two layers each with 576 and 432 dimensions. Three hidden layers are employed for the SNN whose output dimensions are given as 360, 216 and 144. The power control network has two hidden layers having 288 and 144 dimensions.

For comparison, a local optimum solution of \eqref{eq:power} is obtained by the majorization minimization (MM) algorithm \cite{HLeeMM:17} which addresses the non-convex objective function via a sequence of convex approximations. At each iteration, an approximated convex MIP is tackled by the Gurobi solver \cite{Gurobi}. As a result, both the association and power control solutions are jointly computed. The Hungarian scheme carries out an alternating optimization between the association and the power control. The power control solution is optimized using the weighted minimum-mean-square-error (WMMSE) algorithm \cite{WMMSE:11} which yields a local optimum solution for a given association obtained with the Hungarian algorithm.
\begin{figure}[t]
\begin{center}
\includegraphics[width=3.1in]{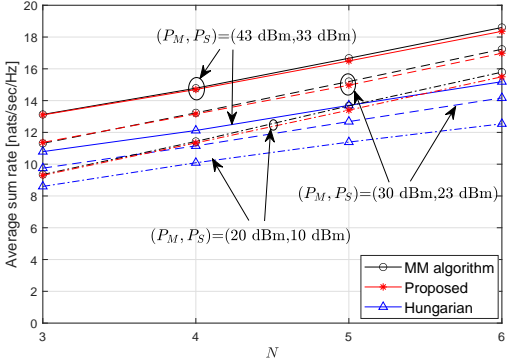}
\end{center}
\vspace{-4mm}
\caption{Average sum-rate performance with respect to $N$.}
\vspace{-6mm}
\label{figure:Cell association}
\end{figure}

Fig. \ref{figure:Cell association} shows the average sum rate performance of various schemes with respect to the number of UEs $N$. Three different scenarios are considered according to the level of the transmit power constraints. In the figure, $(P_{M},P_{S})$ denotes the system with power budgets of $P_{M}$ \& $P_{S}$ at the macro-BS and small-cell BSs, respectively. The performance gain of the proposed SNN approach over the Hungarian algorithm gets larger as the network size $N$ and the power budget $(P_{M},P_{S})$ increase. This implies that the joint optimization of transmit power levels and cell associations becomes significant for a larger MIP with more optimization variables and enlarged feasible spaces. At all simulated $(P_{M},P_{S})$ and $N$, the proposed SNN achieves the almost identical performance to that of the MM algorithm. This verifies the effectiveness of the SNN as a non-convex MIP solver for arbitrary problem size. 



Table II presents the time complexity of various schemes by evaluating the CPU running time. For fair comparison, all schemes are realized by Python with Numpy library. Note that the batch operations are not involved for the implementation of the SNN. It is seen that the proposed SNN approach can significantly reduce the execution time compared to the baseline method while achieving the almost identical performance to the MM algorithm. The execution time of the baseline schemes quickly increases with $N$ and $(P_{M},P_{S})$. In general, more repetitions are required as the size of a problem $N$ and feasible set specified by $P_{M}$ and $P_{S}$ grow. In contrast, the time complexity of the proposed method is independent of the power constraints since the computations of the trained SNN is dominated by its structure, e.g., the number of layers. As discussed before, the SNN would need slightly more calculations as $N$ gets larger. From these results, it is concluded that the proposed approach is a promising non-convex MIP solver which achieves a good trade-off between the performance and the complexity.

\begin{table}[t]
\centering
\caption{Average CPU execution time [sec]}
\vspace{-2mm}
\begin{tabular}{|c||c|c|c|c|}
\hline
    & \multicolumn{2}{c|}{$N=3$} & \multicolumn{2}{c|}{$N=6$} \\
    \cline{1-5}
\hline
$(P_M,P_S)$  &\!\!\!$(20,10)$\!\!\!&\!\!\!$(43,33)$\!\!\!&\!\!\!$(20,10)$\!\!\!&\!\!\!$(43,33)$\!\!\!   \\ \cline{1-5}
\hhline{|=||=|=|=|=|}
MM & $4.37\times10^3$ & $4.89\times10^3$   & $1.43\times10^4$  &  $1.94\times10^4$   \\ \cline{1-5}
\hline
Hungarian & $7.27\times10$ & $1.17\times10^2$  & $1.21\times10^2$ & $2.50\times10^2$ \\ \cline{1-5}
\hline
Proposed & \multicolumn{2}{c|}{5.19}   & \multicolumn{2}{c|}{6.11}  \\ \cline{1-5}
\hline
\end{tabular}
\vspace{-3mm}
\end{table}

\vspace{-3mm}
\section{Conclusions}
\vspace*{-1mm}
This work has studied a DL framework for handling non-convex assignment problems in wireless networks. To address combinatorial binary constraints, an SNN architecture has been proposed which carries out non-convex projections onto permutation matrix spaces. The viability of the proposed DL approach has been demonstrated in various networking scenarios.
\vspace{-3mm}
\bibliographystyle{ieeetr}
\begingroup
\renewcommand{\baselinestretch}{0.92}
\vspace*{-1mm}
\bibliography{AZREF}
\endgroup

\end{document}